\documentclass[letter]{article}
\usepackage{booktabs} 

\usepackage[ruled]{algorithm2e} 

\usepackage[utf8]{inputenc}
\usepackage[english]{babel}

\usepackage{amsmath}
\usepackage{amsthm}
\usepackage{graphics}	
\usepackage{graphicx,psfrag,epsf}
\usepackage{enumerate}
\usepackage{natbib}
\usepackage{amssymb}
\usepackage{setspace}
\usepackage{fancyhdr}
\usepackage{url}

\theoremstyle{plain}
\newtheorem{theorem}{Theorem}[section]

\theoremstyle{definition}
\newtheorem{definition}[theorem]{Definition}

\newcommand{\vect}[1]{\mathbf{#1}}

\begin{document}
\title{A novel Empirical Bayes with Reversible Jump Markov Chain in User-Movie Recommendation system}

\maketitle
\begin{center}
\author{\textbf{Arabin Kumar Dey and Himanshu Jhamb}\\ ~~~\\ Department of Mathematics, IIT Guwahati}
\end{center}


\begin{abstract}
  In this article we select the unknown dimension of the feature by reversible jump MCMC inside a simulated annealing in bayesian set up of collaborative filter. We implement the same in MovieLens small dataset.  We also tune the hyper parameter by using a modified empirical bayes.  It can also be used to guess an initial choice for hyper-parameters in grid search procedure even for the datasets where MCMC oscillates around the true value or takes long time to converge.   
\end{abstract}

%
%

%
%

\noindent%
\vfill


\lhead{}
\rhead{Movie_reco_RJMCMC}

\section{Introduction}

  Matrix factorization is a popular technique in collaborative filtering to predict the missing ratings in user-movie recommendation system.  In a very recent paper of Dey et al. \cite{dey2017case}, bayesian formulation of this problem is discussed.  Hyper prameter choice was the major issue in that paper.  However this problem was attempted only after suitable choice of feature dimension for user and movie feature vector.  Usual method used to select such feature dimension is nothing but one dimensional grid search that select the dimension which minimizes the test error.  This approach is boring as it attracts extra computational burden to select the feature dimension.  In this paper we adapt a bayesian paradigm which automatically select the feature dimension along with hyper parameters which minimizes the cost function.  We propose to use simulated annealing instead of straight forward metropolis hastings so that we can extract the global optimal value of the feature vector.  The idea of using simulated annealing in collaborative filter is recently attempted by few researchers [\cite{feng2013application}, \cite{shehata2017simulated}, \cite{picot2010semantic}].  However selecting the feature dimension automatically in user-movie recommendation set up is not attempted by anyone of the author.  We propose to use reversible jump MCMC in simulated annealing set up.  However such optimization still requires grid search on hyper parameters. So we propose to use empirical bayes used by Atchade to take care the choice of hyper parameters.  But this does not fully solve the implementation problem as such algorithm may depend on initial choice of the hyper-parameters.  So we modify the empirical bayes further and we call it an Adam-based empirical algorithm.        
 
  Bayesian model is used in many papers related to collaborative filter (\cite{chien1999bayesian}, \cite{jin2004bayesian}, \cite{xiong2010temporal}, \cite{yu2002collaborative}).  \cite{ansari2000internet} propose a bayesian preference model that statistically involves several types of information useful for making recommendations, such as user preferences, user and item features and expert evaluations.  A fully bayesian treatment of the Probabilistic Matrix Factorization (PMF) model is discussed by \cite{salakhutdinov2008bayesian} in which model capacity is controlled automatically under Gibbs sampler integrating over all model parameters and hyperparameters.  One of the important objective of Collaborative Filter (\cite{su2009survey}, \cite{liu2009probabilistic}) is to handle the highly sparse data.  Our proposed automation in the algorithm allow us not only to take care such problems, but it also makes the algorithm more scalable.

 The rest of the paper is orgnized as follows.  In section 2, we provide usual collaborative filter algorithm and address few problems.  Implementation of simulated annealing along with Reversible Jump MCMC and Empirical Bayes method is used in section 3.  Results on user-movie data set is provided in Section 4.  We conclude the paper in section 5.

\section{Learning latent features}

\begin{definition}\label{abc1}
$n$ := Number of unique users,\\
$p$ := Number of unique items,\\
$k$ := Number of latent feature,\\
$\vect{M}$ := Sparse rating matrix of dimension $(n \times p)$ where the $(i, j)^{th}$ element of the rating $m_{ij}$ is given by user $i$ to item $j$.\\
$\vect{U}$ := The user feature matrix of dimension $(n \times k)$ where row $i$ represents the user feature vector $\vect{u}_i$.\\
$\vect{V}$ := The item feature matrix of dimension $(p \times k)$ where row $j$ represents the item feature vector $\vect{v}_j$.\\
$\mathcal{L}$ := The loss function which is to be minimized.\\
$\lambda_1$ \& $\lambda_2$ := User and item hyper-parameters in the regularized Loss function $\mathcal{L}$.\\
$|| \cdot ||_F$ := Frobenius norm for matrices.\\
$\kappa$ := The set of user-item indices in the sparse rating matrix $\vect{M}$ for which the ratings are known.\\
$\kappa_i$ := The set of indices of item indices for user $i$ for which the ratings are known.\\
$\kappa_j$ := The set of indices of user indices for item $j$ who rated it.
\end{definition}

 When latent feature dimension $k$ is known,  the system  minimizes the regularized loss on the set of known ratings to find optimal set of parameter vectors $\vect{u}_i$ and $\vect{v}_j$, for all $i$ and $j$.

\begin{equation}\label{eqn1}
\mathcal{L}(\vect{M}; \vect{U},\vect{V},\lambda_1, \lambda_2) = \frac{1}{|\kappa|}\sum_{i,j \in \kappa} (m_{ij} - \vect{u}_i^T.\vect{v}_j)^2 + \lambda_1||\vect{U}||_F^2 + \lambda_2||\vect{V}||_F^2
\end{equation}

  This Loss function is a biconvex function in $\vect{U}$ and $\vect{V}$ and can be iteratively optimized by regularized least square methods keeping the hyper-parameters fixed.

  The user and item feature matrices are first heuristically initialized using normal random matrices with iid. entries such that the product $\vect{U}.\vect{V}^T$ has a mean of 3 and variance 1.\\
The iteration is broken into 2 steps until test loss convergence.
The first step computes the regularized least squares estimate for each of the user feature vectors $\vect{u}_i$ from their known ratings.\\
The second step computes the regularized least squares estimate for each of the item feature vectors $\vect{v}_j$ from their known ratings.\\

The first step minimizes the below expression keeping $\vect{V}$ fixed.
\begin{equation} \label{eqn2}
|| \vect{M}_{i,\kappa_i} - \vect{V}_{\kappa_i}.\vect{u}_i||^2 + \lambda_1||\vect{u}_i||^2 \qquad \forall i = 1, 2, \ldots, n.
\end{equation}

The second step minimizes the below expression keeping $\vect{U}$ fixed.
\begin{equation}\label{eqn3}
|| \vect{M}_{\kappa_j,j} - \vect{U}_{\kappa_j}.\vect{v}_j||^2 + \lambda_2||\vect{v}_j||^2 \qquad \forall j = 1, 2, \ldots, p.
\end{equation}

The normal equations corresponding to the regularized least squares solution for user feature vectors are
\begin{equation}\label{eqn4}
(\vect{V}_{\kappa_i}^T.\vect{V}_{\kappa_i} + \lambda_1 \vect{I}_{k}).\vect{u}_i = \vect{V}_{\kappa_i}^T.\vect{M}_{i,\kappa_i} \qquad i = 1, 2, \ldots, n
\end{equation}
The normal equations corresponding to the regularized least squares solution for item feature vectors are
\begin{equation}\label{eqn4}
(\vect{U}_{\kappa_j}^T.\vect{U}_{\kappa_j} + \lambda_2 \vect{I}_{k}).\vect{v}_1 = \vect{U}_{\kappa_j}^T.\vect{M}_{\kappa_j,j} \qquad i = 1, 2, \ldots, n
\end{equation}

  Iteratively minimizing the loss function gives us a local minima (possibly global).

\subsection{Drawbacks of the previous method}  One of the major drawback of the previous method is finding the choice of $\lambda_1$ and $\lambda_2$. Usual method uses grid search method.  Moreover choice of feature dimension makes the grid search approach computationally less efficient.

\section{Proposed Method}  We propose to use simulated annealing \cite{welsh1989simulated} in this set up which connects this optimization set in bayesian paradigm.  Reseacher has already used this optimization in some recent work.  However capturing the dimension simultaneously in simulated annealing is not available.  We use reversible jump MCMC inside metropolis hastings steps of simulated annealing.  
 
\subsection{Ordinary Simulated Annealing in User-movie set up}  

  Given the loss function in equation-\ref{eqn1}, we get the corresponding Boltzmann Distribution as posterior distribution of $\vect{U}, \vect{V}$ given $\vect{M}$,

\begin{equation}\label{eqn5}
p(\vect{U}, \vect{V} | \vect{M}, \lambda_1, \lambda_2) = e^{-\frac{1}{|\kappa|T}[\sum_{i, j \in \kappa} (m_{ij} - \vect{u}_i^T.\vect{v}_j)^2 + \lambda_1||\vect{U}||_F^2 + \lambda_2||\vect{V}||_F^2]}
\end{equation}

where $T$ is the temperature and each component of prior vectors or user and movie feature vectors are independent and identically distributed normal variates.  Variance of each of the user movie parameters are $\frac{T}{\lambda_1}$ and $\frac{T}{\lambda_{2}}$. Clearly $\lambda_1$ and $\lambda_2$ are hyper parameters.

\subsection{Simulated Annealing with Reversible Jump MCMC in Empirical Bayes Set up}

  Reversible Jump MCMC is used when dimension in the given problem itself is an unknown quantity.  In this context, user and movie feature dimension is unknown. We assume maximum possible dimension as 50.  At each iteration of MCMC, we choose a dimension uniformly from the set $\{ 1, 2, \cdots, 50 \}$.  Temparature is set in simulated annealing as $ T_{i} = \beta T_{i - 1} $.  At each move, new point $Y$ is chosen based on old point $X$ via an orthogonal Gram-Chalier transformation $Y = AX$. If $Y = \{ y_{1}, y_{2}, \cdots, y_{n} \}$ and $X = \{ x_{1}, x_{2}, \cdots, x_{n} \}$, then Gram-Chalier orthogonal transformation uses the following relation :
\begin{eqnarray*}
y_{1} & = & \frac{1}{\sqrt{n}} (x_{1} + x_{2} + \cdots + x_{n});\\
y_{2} & = & \frac{1}{\sqrt{2}} (x_{1} - x_{2});\\
&\vdots& \\
y_{i} &= & \frac{1}{\sqrt{i (i - 1)}} (x_{1} + x_{2} + \cdots + x_{i - 1} -  (i - 1)x_{i});\\
 &\vdots &\\
y_{n} &=& \frac{1}{\sqrt{n (n - 1)}} (x_{1} + x_{2} + \cdots + x_{n - 1} -  (n - 1)x_{n});
\end{eqnarray*}

  Since, choice of hyperparameter $\lambda_1$ and $\lambda_2$ is crucial, we use empirical bayes to select them based on the data set.  In one of the case-study in user-movie set up, Dey et al. \cite{dey2017case} used empirical bayes proposed by Atchade \cite{atchade2011computational}.  The algorithm is as follows : 

The Stochastic approximation algorithm proposed to maximize the function of hyperparameter given the observations ($l(\lambda \mid y)$) as :
\begin{itemize}
  \item Generate $\theta_{n+1} = P_{\lambda_n}(\theta, \cdot)$.
  \item Calculate $\lambda_{n+1} = \lambda_n + a_n H(\lambda_n, \theta_{n+1})$
\end{itemize}
  where, $\theta_n$ is the parameter vector generated from transition kernel $P_{\lambda_n}(\theta, \cdot)$ given the hyper-parameter $\lambda_n$ where $H(\lambda_n, \theta_{n+1})$ is such that
\begin{equation}
 H(\lambda, \theta) := \nabla_{\lambda} \log (f_{\theta, \lambda}(y)\pi(\theta \mid \lambda)) \label{H}
\end{equation}
Note that $f_{\theta, \lambda}(y)\pi(\theta \mid \lambda)$ is the posterior function of the parameter $\theta$ given the data y.

  First step of stochastic approximation algorithm is to draw sample from transition kernel $P_{\theta}$ which is our proposed reversible jump markov chain with simulated annealing algorithm to obtain a sequence of random samples from $\pi(\vect{U}, \vect{V} \mid \vect{M}, \lambda_1, \lambda_2)$.  Let $\vect{U}^{(i)}$ and $\vect{V}^{(i)}$ be the current iterates of the iteration sequence and $q(\vect{U}, \vect{V} \mid \vect{U}^{(i)}, \vect{V}^{(i)})$ be the proposal distribution.  The algorithmic steps to get the next iterate is
\begin{itemize}
        \item Set : Initial choice of $k = k^{0} \sim Unif\{1, 2, \cdots, 50 \}$ 
        \item $\theta_{k^{0}} = (U_{k^{0}}, V_{k^{0}})$, $T_{0}, \lambda_{1}, \lambda_{2}.$
        \item while $T_{0} > T_{min}$, 
             \begin{itemize}
               \item $r \sim Unif(0, 1)$.
               \item $k^{'} \sim$ Unif(0, 1).
               \item if($k^{'} > k^{0}$)
                \item Generate Unif(0, 1) of dimension $2(k^{'} - k^{0})$, which contains $(k^{'} - k^{0})$ numbers from $N(0, \sqrt{\frac{T_{0}}{\lambda_{1}}})$ and $(k^{'} - k^{0})$ numbers from $N(0, \sqrt{\frac{T_{0}}{\lambda_{2}}})$.
               \item $x^{'} = h(x, u) = A \times(x, u)$; where $h(\cdot)$ is a Gram-chalier orthogonal transformation such that $|J| = |\frac{\delta(x, u)}{\delta(x^{'}, u^{'})}| = 1$
               \item Calculate $\alpha(x, x^{'}) = \min\{ 1, \frac{e^{-L_{0}(U^{k^{'}}, V^{k^{'}}; M, \lambda_{1}, \lambda_{2})} J(k^{'} \rightarrow k^{0})}{e^{-L_{0}(U^{k^{0}}, V^{k^{0}}; M, \lambda_{1}, \lambda_{2})} g(u) J(k^{0} \rightarrow k^{'})}\}$         
               where move probability $J(k^{'} \rightarrow k^{0}) = \frac{(k^{'} - 1)}{n}; J(k^{0} \rightarrow k^{'}) = \frac{n - k^{0}}{n}$
               \item if($r <  \alpha(x, x^{'})$)
               \item accept $x^{'}$ and update the hyperparameters using empirical bayes.
               \item else accept x and update hyper-parameters using empirical bayes.   
             \end{itemize}  	
        \item if($k^{'} < k^{0}$) 
        \item Make a birth move from $k^{'}$ to $k^{0}$. We should get $ (x , u) = h^{'}(x') = A^{'} x^{'}$ so that $|J| = 1$, where $h^{'}(\cdot)$ is the inverse map of $h(\cdot)$. 
        \item $\alpha(x, x^{'}) = \min\{ 1, \frac{e^{-L_{0}(U^{k^{'}}, V^{k^{'}}; M, \lambda_{1}, \lambda_{2})} g(u) J(k^{'} \rightarrow k^{0})}{e^{-L_{0}(U^{k^{0}}, V^{k^{0}}; M, \lambda_{1}, \lambda_{2})} J(k^{0} \rightarrow k^{'})}\}$ where, $ J(k^{0} \rightarrow k^{'}) = \frac{k^{0} - 1}{n}$ and $J(k^{0} \rightarrow k^{'}) = \frac{n - k^{'}}{n}$   
        \item if($r <  \alpha(x, x^{'})$)
               \item accept $x^{'}$ and update the hyperparameters using empirical bayes.
               \item else accept x and update hyper-parameters using empirical bayes.     
        \item if($k^{'} = k^{0}$) 
        \item Choose $x^{'}$  from dimension $k^{'}$ from $N(0, 1)$.
        \item Apply usual simulated annealing and update the hyper parameters as similar to previous one.
        \item At each iteration decrease the temparature.
\end{itemize}                       

 According to the algorithm, we sample one set of $\vect{U}$ and $\vect{V}$ using Metropolis-Hasting algorithm and use them to update hyper-parameters in the next step.

\begin{equation}
\lambda_1^{(i+1)} = \lambda_1^{(i)} + a_n H(\lambda_1^{(i)},(\vect{U}^{(i+1)},\vect{V}^{(i+1)}))
\end{equation}
and
\begin{equation}
\lambda_2^{(i+1)} = \lambda_2^{(i)} + a_n H(\lambda_2^{(i)},(\vect{U}^{(i+1)},\vect{V}^{(i+1)}))
\end{equation}
From the definition of $H$ in Equation-\ref{H}, we have
\begin{eqnarray}
H(\lambda_1^{(i)},(\vect{U}^{(i+1)},\vect{V}^{(i+1)})) &=& \nabla_{\lambda_1} \log \pi(\vect{U}^{(i+1)} \mid \lambda_1) \nonumber\\
&=& \nabla_{\lambda_1} \{ -\frac{\lambda_1}{T_{i}} ||\vect{U}^{(i+1)}||_F^2 \} \nonumber\\
&=& -\frac{1}{T_{i}}||\vect{U}^{(i+1)}||_F^2
\end{eqnarray}
similarly
\begin{eqnarray}
H(\lambda_2^{(i)},(\vect{U}^{(i+1)},\vect{V}^{(i+1)})) &=& \nabla_{\lambda_2} \log \pi(\vect{V}^{(i+1)} \mid \lambda_2) \nonumber\\
&=& \nabla_{\lambda_2} \{ -\frac{\lambda_2}{T_{i}} ||\vect{V}^{(i+1)}||_F^2 \} \nonumber\\
&=& -\frac{1}{T_{i}}||\vect{V}^{(i+1)}||_F^2
\end{eqnarray}

 In this case the sequence $\{a_n,  n > 0 \}$ is chosen adaptively based on a well-known algorithm called Adam.  


  Adam needs the following steps of tuning the hyper parameters in context of above empirical bayes set up.

\begin{itemize}

\item  The following hyper parameters are required :
$\alpha$ = step size; $\beta_1, \beta_2 \in [0, 1) :$ Exponential decay rates for moment estimates. $f(\theta)$ : stochastic objective function with parameter $\theta$.  $\theta_{0} = (\lambda_1, \lambda_2)$ Initial choice of parameter vector. 

\item Assume $\theta_{0}$ as initial choice of parameter vector. Initialize $m_{0} = 0$, $\nu_{0} = 0$ and $t = 0$.

\item Repeat the following steps until convergence.

\begin{itemize}

\item $ \underline{m}_{i + 1, 1} = \beta_{1} \underline{m}_{i, 1} + (1 - \beta_{1}) H(\lambda_1^{(i)},(\vect{U}^{(i+1)},\vect{V}^{(i+1)})).$ [Update biased first moment e stimate]

\item $ \underline{m}_{i + 1, 2} = \beta_{1} \underline{m}_{i, 2} + (1 - \beta_{1}) H(\lambda_2^{(i)},(\vect{U}^{(i+1)},\vect{V}^{(i+1)})).$ [Update biased first moment estimate]

\item $ \underline{\nu}_{i + 1, 1} = \beta_{2} \underline{m}_{i, 2} + (1 - \beta_{2}) (H(\lambda_1^{(i)},(\vect{U}^{(i+1)},\vect{V}^{(i+1)})))^2.$ [Update biased first moment estimate]

\item $ \underline{\nu}_{i + 1, 2} = \beta_{2} \underline{\nu}_{i, 2} + (1 - \beta_{2}) (H(\lambda_2^{(i)},(\vect{U}^{(i+1)},\vect{V}^{(i+1)})))^2.$ [Update second raw moment estimate]

\item $ \underline{m}_{i + 1, 1} = \frac{\underline{m}_{i + 1, 1}}{1 - \beta_{1}}$ ; $ \underline{m}_{i + 1, 2} = \frac{\underline{m}_{i + 1, 2}}{1 - \beta_{1}}$ ; [Compute bias-corrected 1st moment] $\underline{\nu_{i + 1, 1}} = \frac{\underline{\nu_{i + 1, 1}}}{1 - \beta_{2}}$, $\underline{\nu_{i + 1, 2}} = \frac{\underline{\nu_{i + 1, 2}}}{1 - \beta_{2}}$ [Compute bias-corrected 2nd moment]

\item $\underline{\lambda}_{i + 1, 1} = \underline{\lambda}_{i + 1, 1} - \alpha \frac{\underline{\hat{m}}_{i + 1, 1}}{\sqrt{\underline{\hat{\nu}}_{i + 1, 1} + \epsilon}}$ $\underline{\lambda}_{i + 1, 2} = \underline{\lambda}_{i + 1, 2} - \alpha \frac{\underline{\hat{m}}_{i + 1, 1}}{\sqrt{\underline{\hat{\nu}}_{i + 1, 1} + \epsilon}}$

\item $i = i + 1$; 

\end{itemize}

\item Return $\underline{\lambda}_{i + 1, 1}$, $\underline{\lambda}_{i + 1, 2}$.

\end{itemize}

\section{Data Analysis and Results} The benchmark datasets used in the experiments were the 3 Movie Lens dataset.  They can be found at \url{http://grouplens.org/datasets/movielens/}.  We implement the above algorithm in old MovieLens small dataset which has been used with 943 users and 1682 movie. We take 80\% data as training and 20\% data as test set.  Initial choice of hyparameters is set to $\lambda_1 = 30$ and $\lambda_2 = 30$, $\beta_1 = 0.9$, $\beta_2 = 0.999$, $\epsilon = 1e-08$, $\alpha = 0.001$.  The vectors $\vect{U}_0$ and $\vect{V}_0$ were initialized with iid standard normal random numbers.  We stop the updates of Hyper-Parameters $\lambda_1$ and $\lambda_2$ if changes in succesive two hyper-parameters falls below tolerance, where tolerance is set to $10^{-5}$.  

\begin{figure}[h!]
  \includegraphics[width = 0.5\textwidth]{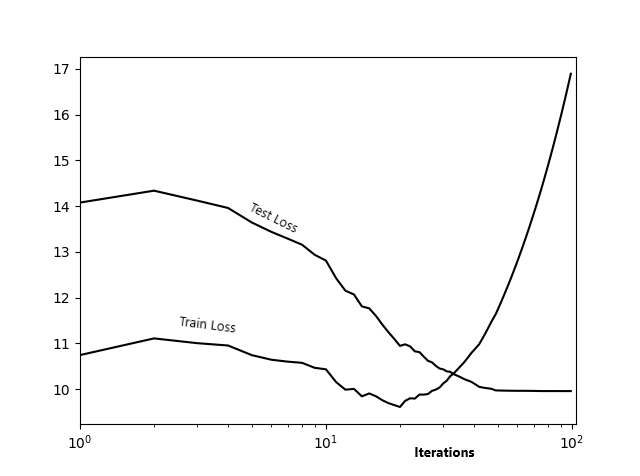}
  \caption{Test and Train smoothed Loss over different iteration}
  \label{fig:one}
\end{figure}

  Test and Train smoothed Loss is drawn based on a fixed size window of 10 iterations in Figure-\ref{fig:one}.  Minimum Training Cost can be found at k=2 and Test RMSE at this point as: 1.02089331071.  However the test loss stabilizes at k = 4.  Converged Hyper-Parameters: 19.439, 25.796.    

\section{Conclusion}  The paper describes successful implementation of choice of hidden feature dimension along with automatically tuning all hyper-parameters based on the data set in a  collaborative filter set up.  The result depends on the selection of test and trainng set.  Transformation matrix calculation makes the algorithm little slow. We can modify the algorithm taking levy-type proposal density in this automated simulated annealing.  There some variation of the model can proposed in this direction.  The work is on progress. 

\bibliographystyle{chicago}
\bibliography{rec-bibliography-biblatex}


\begin{thebibliography}{14}


\ifx \showCODEN    \undefined \def \showCODEN     #1{\unskip}     \fi
\ifx \showDOI      \undefined \def \showDOI       #1{#1}\fi
\ifx \showISBNx    \undefined \def \showISBNx     #1{\unskip}     \fi
\ifx \showISBNxiii \undefined \def \showISBNxiii  #1{\unskip}     \fi
\ifx \showISSN     \undefined \def \showISSN      #1{\unskip}     \fi
\ifx \showLCCN     \undefined \def \showLCCN      #1{\unskip}     \fi
\ifx \shownote     \undefined \def \shownote      #1{#1}          \fi
\ifx \showarticletitle \undefined \def \showarticletitle #1{#1}   \fi
\ifx \showURL      \undefined \def \showURL       {\relax}        \fi
\providecommand\bibfield[2]{#2}
\providecommand\bibinfo[2]{#2}
\providecommand\natexlab[1]{#1}
\providecommand\showeprint[2][]{arXiv:#2}

\bibitem[\protect\citeauthoryear{Ansari, Essegaier, and Kohli}{Ansari
  et~al\mbox{.}}{2000}]%
        {ansari2000internet}
\bibfield{author}{\bibinfo{person}{Asim Ansari}, \bibinfo{person}{Skander
  Essegaier}, {and} \bibinfo{person}{Rajeev Kohli}.}
  \bibinfo{year}{2000}\natexlab{}.
\newblock \bibinfo{title}{Internet recommendation systems}.
\newblock   (\bibinfo{year}{2000}).
\newblock


\bibitem[\protect\citeauthoryear{Atchad{\'e}}{Atchad{\'e}}{2011}]%
        {atchade2011computational}
\bibfield{author}{\bibinfo{person}{Yves~F Atchad{\'e}}.}
  \bibinfo{year}{2011}\natexlab{}.
\newblock \showarticletitle{A computational framework for empirical Bayes
  inference}.
\newblock \bibinfo{journal}{\emph{Statistics and computing}}
  \bibinfo{volume}{21}, \bibinfo{number}{4} (\bibinfo{year}{2011}),
  \bibinfo{pages}{463--473}.
\newblock


\bibitem[\protect\citeauthoryear{Chien and George}{Chien and George}{1999}]%
        {chien1999bayesian}
\bibfield{author}{\bibinfo{person}{Yung-Hsin Chien} {and}
  \bibinfo{person}{Edward~I George}.} \bibinfo{year}{1999}\natexlab{}.
\newblock \showarticletitle{A bayesian model for collaborative filtering}. In
  \bibinfo{booktitle}{\emph{AISTATS}}.
\newblock


\bibitem[\protect\citeauthoryear{Dey, Somani, and Acharyya}{Dey
  et~al\mbox{.}}{2017}]%
        {dey2017case}
\bibfield{author}{\bibinfo{person}{Arabin~Kumar Dey}, \bibinfo{person}{Raghav
  Somani}, {and} \bibinfo{person}{Sreangsu Acharyya}.}
  \bibinfo{year}{2017}\natexlab{}.
\newblock \showarticletitle{A case study of empirical Bayes in a user-movie
  recommendation system}.
\newblock \bibinfo{journal}{\emph{Communication in Statistics : Case Studies,
  Data Analysis and Applications}} \bibinfo{volume}{3}, \bibinfo{number}{1 -2}
  (\bibinfo{year}{2017}), \bibinfo{pages}{1--6}.
\newblock


\bibitem[\protect\citeauthoryear{Feng and Su}{Feng and Su}{2013}]%
        {feng2013application}
\bibfield{author}{\bibinfo{person}{Zhi~Ming Feng} {and} \bibinfo{person}{Yi~Dan
  Su}.} \bibinfo{year}{2013}\natexlab{}.
\newblock \showarticletitle{Application of Using Simulated Annealing to Combine
  Clustering with Collaborative Filtering for Item Recommendation}. In
  \bibinfo{booktitle}{\emph{Applied Mechanics and Materials}},
  Vol.~\bibinfo{volume}{347}. Trans Tech Publ, \bibinfo{pages}{2747--2751}.
\newblock


\bibitem[\protect\citeauthoryear{Jin and Si}{Jin and Si}{2004}]%
        {jin2004bayesian}
\bibfield{author}{\bibinfo{person}{Rong Jin} {and} \bibinfo{person}{Luo Si}.}
  \bibinfo{year}{2004}\natexlab{}.
\newblock \showarticletitle{A Bayesian approach toward active learning for
  collaborative filtering}. In \bibinfo{booktitle}{\emph{Proceedings of the
  20th conference on Uncertainty in artificial intelligence}}. AUAI Press,
  \bibinfo{pages}{278--285}.
\newblock


\bibitem[\protect\citeauthoryear{N.~Liu, Zhao, and Yang}{N.~Liu
  et~al\mbox{.}}{2009}]%
        {liu2009probabilistic}
\bibfield{author}{\bibinfo{person}{Nathan N.~Liu}, \bibinfo{person}{Min Zhao},
  {and} \bibinfo{person}{Qiang Yang}.} \bibinfo{year}{2009}\natexlab{}.
\newblock \showarticletitle{Probabilistic latent Preference Analysis for
  Collaborative Filtering}. In \bibinfo{booktitle}{\emph{Proceedings of the
  18th ACM conference on information and knowledge management}}. ACM,
  \bibinfo{pages}{759--766}.
\newblock


\bibitem[\protect\citeauthoryear{Picot-Clemente, Cruz, and
  Nicolle}{Picot-Clemente et~al\mbox{.}}{2010}]%
        {picot2010semantic}
\bibfield{author}{\bibinfo{person}{Romain Picot-Clemente},
  \bibinfo{person}{Christophe Cruz}, {and} \bibinfo{person}{Christophe
  Nicolle}.} \bibinfo{year}{2010}\natexlab{}.
\newblock \showarticletitle{A Semantic-based Recommender System Using A
  Simulated Annealing Algorithm}. In \bibinfo{booktitle}{\emph{SEMAPRO 2010,
  The Fourth international Conference on Advances in Semantic Processing}}.
  \bibinfo{pages}{ISBN--978}.
\newblock


\bibitem[\protect\citeauthoryear{Salakhutdinov and Mnih}{Salakhutdinov and
  Mnih}{2008}]%
        {salakhutdinov2008bayesian}
\bibfield{author}{\bibinfo{person}{Ruslan Salakhutdinov} {and}
  \bibinfo{person}{Andriy Mnih}.} \bibinfo{year}{2008}\natexlab{}.
\newblock \showarticletitle{Bayesian probabilistic matrix factorization using
  Markov chain Monte Carlo}. In \bibinfo{booktitle}{\emph{Proceedings of the
  25th international conference on Machine learning}}. ACM,
  \bibinfo{pages}{880--887}.
\newblock


\bibitem[\protect\citeauthoryear{Shehata, Nassef, and Badr}{Shehata
  et~al\mbox{.}}{2017}]%
        {shehata2017simulated}
\bibfield{author}{\bibinfo{person}{Mostafa~A Shehata},
  \bibinfo{person}{Mohammad Nassef}, {and} \bibinfo{person}{Amr~A Badr}.}
  \bibinfo{year}{2017}\natexlab{}.
\newblock \showarticletitle{Simulated Annealing with Levy Distribution for Fast
  Matrix Factorization-Based Collaborative Filtering}.
\newblock \bibinfo{journal}{\emph{arXiv preprint arXiv:1708.02867}}
  (\bibinfo{year}{2017}).
\newblock


\bibitem[\protect\citeauthoryear{Su and Khoshgoftaar}{Su and
  Khoshgoftaar}{2009}]%
        {su2009survey}
\bibfield{author}{\bibinfo{person}{Xiaoyuan Su} {and} \bibinfo{person}{Taghi~M
  Khoshgoftaar}.} \bibinfo{year}{2009}\natexlab{}.
\newblock \showarticletitle{A survey of collaborative filtering techniques}.
\newblock \bibinfo{journal}{\emph{Advances in artificial intelligence}}
  \bibinfo{volume}{2009} (\bibinfo{year}{2009}), \bibinfo{pages}{4}.
\newblock


\bibitem[\protect\citeauthoryear{Welsh}{Welsh}{1989}]%
        {welsh1989simulated}
\bibfield{author}{\bibinfo{person}{DJA Welsh}.}
  \bibinfo{year}{1989}\natexlab{}.
\newblock \bibinfo{title}{SIMULATED ANNEALING: THEORY AND APPLICATIONS}.
\newblock   (\bibinfo{year}{1989}).
\newblock


\bibitem[\protect\citeauthoryear{Xiong, Chen, Huang, Schneider, and
  Carbonell}{Xiong et~al\mbox{.}}{2010}]%
        {xiong2010temporal}
\bibfield{author}{\bibinfo{person}{Liang Xiong}, \bibinfo{person}{Xi Chen},
  \bibinfo{person}{Tzu-Kuo Huang}, \bibinfo{person}{Jeff Schneider}, {and}
  \bibinfo{person}{Jaime~G Carbonell}.} \bibinfo{year}{2010}\natexlab{}.
\newblock \showarticletitle{Temporal collaborative filtering with bayesian
  probabilistic tensor factorization}. In \bibinfo{booktitle}{\emph{Proceedings
  of the 2010 SIAM International Conference on Data Mining}}. SIAM,
  \bibinfo{pages}{211--222}.
\newblock


\bibitem[\protect\citeauthoryear{Yu, Schwaighoferi, and Tresp}{Yu
  et~al\mbox{.}}{2002}]%
        {yu2002collaborative}
\bibfield{author}{\bibinfo{person}{Kai Yu}, \bibinfo{person}{Anton
  Schwaighoferi}, {and} \bibinfo{person}{Volker Tresp}.}
  \bibinfo{year}{2002}\natexlab{}.
\newblock \showarticletitle{Collaborative ensemble learning : Combining
  collaborative and content based information filtering via hierarchical
  Bayes}. In \bibinfo{booktitle}{\emph{Proceedings of Nineteenth Conference on
  Uncertainty in Artificial Intelligence}}. Morgan Kaufmann Publishers Inc.,
  \bibinfo{pages}{616--623}.
\newblock


\end{thebibliography}


\end{document}